# Critical Survey of the Freely Available Arabic Corpora

**Wajdi Zaghouani**
Carnegie Mellon University Qatar
Computer Science
E-mail: wajdiz@cmu.edu

**Abstract**

The availability of corpora is a major factor in building natural language processing applications. However, the costs of acquiring corpora can prevent some researchers from going further in their endeavours. The ease of access to freely available corpora is urgent needed in the NLP research community especially for language such as Arabic. Currently, there is not easy was to access to a comprehensive and updated list of freely available Arabic corpora. We present in this paper, the results of a recent survey conducted to identify the list of the freely available Arabic corpora and language resources. Our preliminary results showed an initial list of 66 sources. We presents our findings in the various categories studied and we provided the direct links to get the data when possible.

**Keywords:** Arabic, Open source, Free, Corpora, Corpus, Survey.

## 1. Introduction

The use of corpora has been a major factor in the recent advance in natural language processing development and evaluation. However, the high costs of building or licensing a corpora could be an obstacle for many young researchers or even some institution in several parts of the world. Therefore, having access to freely available corpora is clearly a desirable goal. Unfortunately, the freely available corpora are generally not easily found and most resources available from language data providers are for fees or exclusively reserved for subscribers, such as the corpora available from the Linguistic Data Consortium or the Evaluations and Language resources Distribution Agency (ELDA). A simple query for Arabic corpora available in the LDC Catalog shows the availability of 116 corpora of various types (text, speech, evaluation etc…).[1] Another similar query done with the ELRA Corpora search engine showed the availability of 80 corpora.[2] For instance, Arabic can still be considered a relatively resource poor language when compared to other languages such as English, and having access to freely available corpora will definitely improve the current Arabic NLP technologies. In this paper, we present the results of an online survey of the freely available Arabic corpora.

## 2. Current situation of the freely available Arabic corpora

Before starting our survey experiment, we tried various online queries to locate any freely available Arabic corpora or a repository listing the corpora for any easy access to the resources. We found that the information is scattered in various personal and research groups sites that are often not complete or outdated.

As of 2010, ELRA created the LRE Map (Language Resources and Evaluation) which is an online database on language resources. The goal behind LRE Map is to monitor the creation and the use and of language resources. The information is collected during the submission process to LREC and other conferences. We did a query to list the freely available Arabic corpora and we found a limited number and no URL to link the user or project details were available. Habash (2010) listed in his book, various available Arabic corpora sorted by corpus type (speech, text, bi-lingual). Again the list is not designed for the freely available resources and most of data listed are available from data providers. The Association for Computational Linguistics (ACL) maintains a wiki page that lists the available resources by language, the Arabic page only lists five corpora, four free corpora and one proprietary corpora.[3]

The European Network of Excellence in Human Language Technologies (ELSNET) maintains a list of pointers to Arabic and other Semitic NLP and Speech sites, the Arabic resources section includes 23 entries and most of them were created more than 12 years ago.[4]

The Mediterranean Arabic Language and Speech Technology (MEDAR) conducted a survey in 2009 and 2010 to list the existing institutions and experts involved in the development of Arabic language resources, activities and projects being carried out and related and tools.[5] The collected results were compiled and made available into a knowledge base that is accessible online[6]. Again, despite the huge effort made, the list is no longer updated and it lacks the necessary information to locate the data such as the download page or the project description and URL. Finally, we cite other interesting personal efforts to list some Arabic language resources such as the Sibawayh Repository for Arabic Language Processing page,[7] Al-Sulaiti Arabic corpora page[8] and Al-Ghamidi Arabic links page.[9] We consider our efforts described in this project as a complement to what exists already with a focus on the free resources and how each corpus can be obtained.

---

[1] Query performed on January 31st 2014 <http://catalog.ldc.upenn.edu/>

[2] Query performed on January 31st 2014 <http://catalog.elra.info/search.php>

[3] http://aclweb.org/aclwiki/index.php?title=Resources_for_Arabic

[4] http://www.elsnet.org/arabiclist.html

[5] http://www.medar.info/MEDAR_Survey_III.pdf

[6] www.elda.org/medar_knowledge_base/

## 3. The Survey

In order to start the collection of our freely available Arabic corpora list, we created an online survey[10] that was shared in the various NLP related lists such as Corpora[11] and ArabicList.[12] The online survey was intended to be completed within 5-10 minutes to encourage participants, and included some very basic questions such as the provider information, corpus type, size, purpose, download link, related publications, Arabic variety, production status and a confirmation that the corpus is completely free for a research purposes. The online survey was completed by 20 participants who pointed 26 resources. Once the survey results were compiled, we added manually, 40 other freely available Arabic resources taken from the various online sources described in section 2. We also added the missing information when needed (wrong download URL, description, corpus size, authors, etc.). Finally we tried to locate any related publication to the source so it can be cited properly when used. In the next section, we describe briefly a selection of the 66 free resources found during our survey.

## 4. Available Resources

In this section, we present the result of our survey of the freely available Arabic corpora with a focus on the most important work for each of the following categories:

- **Raw Text Corpora**: monolingual corpora, multilingual corpora, dialectal Corpora, web-based corpora.
- **Annotated Corpora:** named entities, error annotation, POS, syntax, semantic, anaphora.
- **Lexicon:** lexical databases and words lists.
- **Speech Corpora:** audio recording, transcribed data.
- **Handwriting Recognition Corpora:** scanned and annotated documents.
- **Miscellaneous Corpora types:** Questions/Answers, comparable corpora, plagiarism detection and summaries.

For each of the four categories, some basic information will be provided a table for each category. It includes the author name or the research group, the corpus name, the corpus size in words or in files. In case there is a publication associated with the corpus, it will be cited as part of the author name in the table otherwise only the corpus download/access URL is provided as a footnote following the corpus name in each table. The sources are sorted in the tables according to their size for the most important to the least important.

### 4.1 Raw Text Corpora

In this section we cite 23 freely available raw text corpora, that is, they do not include any kind of annotation and limited to the text files themselves. The raw text corpora are divided into four categories listed below.

#### 4.1.1. Monolingual Corpora

The 11 freely available monolingual corpora found are all available for download (Table 1).

| Source | Corpus | Words |
|---|---|---|
| Abdelali | Ajdir Corpora[13] | 113,000,000 |
| Alrabiah | KSU Corpus of Classical Arabic[14] | 50,000,000 |
| Saad and Ashour (2010) | OSAC[15] | 18,183,511 |
| Abbas | Alwatan[16] | 10,000,000 |
| Zarrouki | Tashkeela[17] | 6,149,726 |
| Abbas | Al Khaleej[18] | 3,000,000 |
| Al-Thubaity et al. | KACST Arabic Newspaper Corpus[19] | 2,000,000 |
| Al-Saadi | Arabic Words Corpora[20] | 1,500,000 |
| Al-Suleiti | Corpus of Contemporary Arabic[21] | 842,684 |
| Alkanhal et al. (2012) | CRI KACST Arabic Corpus[22] | 235,000 |
| Farwaneh | Arabic Learners Written Corpus[23] | 50,000 |

Table 1: Monolingual Corpora List.

Most of them cover the news domain and they are large size corpora ranging from 1 million words to 113 million words. Other corpora cover other domains such as student essays (Farawaneh) and classical Arabic (KSU Corpus of classical Arabic and Tashkeela). When it comes to data format, we noticed that most of these corpora are stored in text or xml format while others like the Arabic learners corpus, are stored in an inconvenient PDF format that makes it hard to be used for any NLP task.

---

7 http://arab.univ.ma/web/s/
8 http://www.comp.leeds.ac.uk/eric/latifa/arabic_corpora.htm
9 http://www.mghamdi.com/links.htm
10 https://docs.google.com/forms/d/1N2W76d8Uxnzx--0Dj6An2mJr8KzeR0U1rF6pOj6Djjg/viewform?edit_requested=true
11 http://www.hit.uib.no/corpora/
12 https://listserv.byu.edu/cgi-bin/wa?A0=ARABIC-L

13 http://aracorpus.e3rab.com/argistestsrv.nmsu.edu/AraCorpus/
14 http://ksucorpus.ksu.edu.sa/?p=43
15 https://sites.google.com/site/motazsite/arabic/osac
16 http://sourceforge.net/projects/arabiccorpus/
17 http://sourceforge.net/projects/tashkeela/
18 http://sourceforge.net/projects/arabiccorpus/
19 http://sourceforge.net/projects/kacst-acptool/files/?source=navbar
20 http://sourceforge.net/projects/arabicwordcorpu/files/
21 http://www.comp.leeds.ac.uk/eric/latifa/research.htm
22 http://cri.kacst.edu.sa/Resources/TRN_DB.rar
23 http://l2arabiccorpus.cercll.arizona.edu/?q=allFiles

### 4.1.2. Multilingual Corpora

Among the corpora listed in Table 2, we can consider the UN corpus as the most important and the most widely known free corpus for its category. The Meedan with 1 million words Arabic/English aligned sentences, is also a very valuable resource. The Hadith standard corpus and the Quranic Arabic/English aligned corpus included in the Egypt translation tool are less known resources that could be used in any work related to the religious domain.

| Source | Corpus | Words |
|---|---|---|
| Rafalovitch and Dale (2009) | UN Corpus(Arabic portion)[24] | 2,721,463 |
| Bounhas | Hadith Standard Corpus[25] | 2,500,000 |
| Meedan | MEEDAN Translation Memory[26] | 1,000,000 |
| CLSP/JHU | EGYPT Translation Toolkit[27] | 80,000 |

Table 2: Multilingual Corpora List.

### 4.1.3. Dialectal Corpora

The two dialectal corpora listed in the Table 3 below are very valuable, especially that work Arabic dialect processing is a rather recent task there is a real need for such resources. The Tunisian Dialect Corpus (Graja et al.), is a transcribed spoken dialogue corpus formed of 1465 railway staff utterances and 1615 client utterances. The recent work done by (Almeman and Lee 2013), can be considered a major contribution to the advance in the Arabic dialectal resources with its 2 million unique words collected online from 55k webpages and covering four major Arabic dialects (Gulf, Levantine, North Africa, Egypt).

| Source | Corpus | Words |
|---|---|---|
| Almeman and Lee (2013) | Arabic Multi Dialect Text Corpora[28] | 2,000,000 |
| Graja et al. (2010) | Tunisian Dialect Corpus (TuDiCoI)[29] | 3,403 |

Table 3: Caption.

### 4.1.4. Web-based Corpora

In this category we placed some corpora (Table 4) that are exclusively available online through an online query interface so there is no data provided for download which can be inconvenient for some research studies, nevertheless these web-based corpora can be very valuable for concordance and frequency studies given the variety and large size of these corpora (732M words KACST corpus, 317M M words for Leeds and 100M words for ICA and Parkinson corpus), moreover the Arabic variety and text genre covered is large which makes these corpora very suitable for all types of Arabic linguistics studies (Quranic Arabic, classic Arabic, newswire, books etc.).

| Source | Corpus | Words |
|---|---|---|
| Al-Thubaity | KACST Arabic Corpus[30] | 732,780,509 |
| Leeds | Leeds Arabic Internet Corpus[31] | 317,000,000 |
| Alansary et al. (2007) | International Corpus of Arabic[32] | 100,000,000 |
| Parkinson | ArabiCorpus[33] | 100,000,000 |
| Abbas N. | QURANY[34] | 78,000 |
| Sharaf et al. | Quranic Text mining Dataset[35] | 24,000 |

Table 4: Web-based Corpora List.

## 4.2 Annotated Corpora

Annotated corpora are very useful to build systems and tools based on supervised algorithms and the free availability of resources will help young researches to train and build systems at minimal cost. In this section, we list a selection of freely available named entities corpora, Error annotated corpora and some various annotated corpora including part of speech (POS) annotated corpora, syntactically and semantically annotated corpora.

### 4.2.1. Named Entity Corpora

Table 5 lists some very useful resources for the named entities recognition task.

| Source | Corpus | Words |
|---|---|---|
| Steinberger et al. (2011) | JRC-Names[36] | 230,000 |
| Ben Ajiba et al. (2007) | ANERCorp[37] | 150,000 |
| Mohit et al. (2012) | AQMAR Named Entity Corpus[38] | 74,000 |
| Azab et al. (2013) | Named Entity Translation Lexicon[39] | 55,000 |
| Attia et al. 2010 | Named Entities List[40] | 45,202 |
| Ben Ajiba et al. (2007) | ANERGazet[41] | 14,000 |

Table 5: Named Entities Corpora List.

Most of these corpora were reported by their respective authors in major NLP conferences which adds visibility to these resources. The annotation format of these data follows the XML annotation standards put by major evaluation campaigns such as the automatic content extraction (ACE) evaluation campaign.[42] Most of the entries in these resources covers person's names, some organisations and geographical locations names and the size of these data is important, ranging from 14k to 230k.

---

24 http://www.uncorpora.org/
25 http://www.kunuz/
26 https://github.com/anastaw/Meedan-Memory
27 http://old-site.clsp.jhu.edu/ws99/projects/mt/toolkit/
28 http://www.cs.bham.ac.uk/~kaa846/arabic-multi-dialect-text-corpora.html
29 https://sites.google.com/site/anlprg/outils-et-corpus-realises/TuDiCoIV1.xml?attredirects=0

30 http://www.kacstac.org.sa/
31 http://smlc09.leeds.ac.uk/query-ar.html
32 http://www.bibalex.org/ica/en/About.aspx
33 http://arabicorpus.byu.edu/
34 http://quranytopics.appspot.com/
35 http://textminingthequran.com/wiki/Main_Page
36 http://ipsc.jrc.ec.europa.eu/index.php?id=42#c2696
37 http://www1.ccls.columbia.edu/~ybenajiba/downloads.html
38 http://www.ark.cs.cmu.edu/ArabicNER/
39 http://nlp.qatar.cmu.edu/resources/NETLexicon/
40 https://sourceforge.net/projects/arabicnes/
41 http://www1.ccls.columbia.edu/~ybenajiba/downloads.html

### 4.2.2. Error-Annotated Corpora

Error annotated corpora can be very useful for corpus based studies of errors and also for building automatic spelling correction tools. Table 6 lists three resources, QALB and the Arabic learner corpus are still an on-going efforts. The KACST Error corpus (Alkanhal et al. 2012) includes exclusively student essays that are manually corrected while Alfifi et al. (2013) will include the correction of the errors as well as the categories of the errors. When ready, Zaghouani et al. (2014) corpus will be the only 2M words corrected corpus available for Arabic with four text varieties: native and non-native students essays, online users posts and English/Arabic machine translation corrected output.

| Source | Corpus | Words |
|---|---|---|
| Habash et al. (2013) | Qatar Arabic language Bank(QALB)[43] | 2,000,000 |
| Alfifi et al. (2013) | Arabic Learner Corpus[44] | 282,000 |
| Alkanhal et al. (2012) | KACST Error Corrected Corpus[45] | 65,000 |

Table 6: Errors Annotated Corpora List

### 4.2.3. Miscellaneous Annotated Corpora

The corpora listed on Table 7 includes various annotated corpora ranging from semantically annotated corpora to syntactically and morphologically annotated corpora. Most of these resources allows direct download except for OntoNotes that can be obtained freely from the LDC.

The OntoNotes corpus (Weischedel et al. 2013) includes various genres of text (news, conversational telephone speech, weblogs, usenet newsgroups, broadcast, talk shows) in three languages (English, Chinese, and Arabic) with structural information (syntax and predicate argument structure) and shallow semantics (word sense linked to an ontology and coreference).

One notable effort in this category is the ongoing work to build the Quranic Arabic Corpus, an annotated linguistic resource consisting of 77,430 words of Quranic Arabic. The project aims to provide morphological and syntactic annotations for researchers wanting to study the language of the Quran. While the POS annotated version is already available for download, the treebank version is still ongoing. Moreover, an online query interface is available for morphological queries and concordance. Despite its small size, the Arabic Wikipedia dependency corpus is one of the rare freely available Arabic Treebanks.

| Source | Corpus | Words | Type |
|---|---|---|---|
| Weischedel et al. (2013) | OntoNotes Release 5.0[46] | 300,000 | Semantic |
| Dukes and Habash (2010) | The Quranic Arabic Corpus[47] | 77,430 | POS/Syntax |
| Schneider et al. (2012) | AQMAR Arabic Wiki. Supersense Corpus[48] | 65,000 | Semantic |
| Khoja et al. (2001) | Khoja POS tagged corpus[49] | 51,700 | POS |
| E. Mohammed | Arabic Wikipedia Dependency Corpus[50] | 36,000 | Syntax |
| Mezghani et al. (2009) | AnATAr Corpus[51] | 18,895 | Anaphora |

Table 7: Miscellaneous Annotated Corpora List.

## 4.3 Lexicon

In this section we describe some available lexical databases and words lists. Most of these resources are available for download, some of the lexicon are part of tools and systems, but since these tools are open source, these lexicons can be used for research purposes.

### 4.3.1. Lexical Databases

Several efforts have been made in recent years to build various lexical resources for Arabic (Table 8). Fortunately, most of them are free such as the version 1.0 of the well-known Buckwalter morphological analyzer (Buckwalter 2002). Other important efforts were adapted from the English to the Arabic such as the Arabic WordNet (Elkateb et al. 2006) and the Arabic VerbNet (Mousser 2010).

In the Arabic WordNet, the words are grouped into sets of synonyms and it provides general definitions and the various semantic relations between the synonyms sets.
The Arabic VerbNet provides a lexicon in which the most used Arabic verbs are classified and their syntactic and semantic information are provided. An online interface is provided.

### 4.3.2. Words Lists

Table 9 lists various words lists created mostly by Mohammed Attia.[52] These words lists can be used by lexicographers to study various aspects of the Arabic language such as the Arabic MSA word count list. These lists can also be integrated with the lexicons of systems and tools to improve their performances. For instance, the Arabic wordlist of 9M words and the 18k Arabic unknown words list, can be used in a the spell checking systems. Furthermore, the Arabic stop words list of 13k can be used in various application as word filter list.

---

42 http://www.itl.nist.gov/iad/894.01/tests/ace/

43 http://nlp.qatar.cmu.edu/qalb/

44 http://www.comp.leeds.ac.uk/scayga/alc/corpus%20files.html

45 http://cri.kacst.edu.sa/Resources/TST_DB.rar

46 http://catalog.ldc.upenn.edu/LDC2013T19

47 http://corpus.quran.com/download/

48 http://www.ark.cs.cmu.edu/ArabicSST/

49 http://zeus.cs.pacificu.edu/shereen/research.htm#corpora and email the author

50 http://www.ark.cs.cmu.edu/ArabicDeps/

51 https://sites.google.com/site/anlprg/outils-et-corpus-realises/AnATArcorpus-BEB.rar?attredirects=0

52 http://www.attiaspace.com/

| Source | Corpus | Words |
|---|---|---|
| Buckwalter | BAMA 1.0 English-Arabic Lexicon[53] | 82,158 |
| Salmone | Arabic-English Learner's Dictionary[54] | 74,000 |
| Doumi et al. (2013) | Unitex Arabic Package[55] | 50,407 |
| Boudelaa and Wilson (2010) | ARALEX Online[56] | 37,494 |
| Attia et al. (2011) | AraComLex Arabic Lexical Database[57] | 30,000 |
| Mousser (2010) | Arabic VerbNEt[58] | 23,341 |
| Elkateb et al. (2006) | Arabic WordNet[59] | 18,957 |
| Mesfar and Silberztein (2008) | NOOJ Arabic Dictionary[60] | 10,000 |
| ArabEyes | Qamoose[61] | N.A |

Table 8: Lexical Databases List.

| Source | Corpus | Words |
|---|---|---|
| Attia et al. (2011) | Word Count of Modern Standard Arabic[62] | 1,000,000,000 |
| Attia et al. (2012a) | Arabic Wordlist for Spellchecking[63] | 9,000,000 |
| Attia et al. (2010) | Multiword Expressions[64] | 34,658 |
| Attia et al. 2012b | Arabic Unknown Words[65] | 18,000 |
| Zarrouki | Arabic Stop words[66] | 13,000 |
| Attia et al. 2011b | Obsolete Arabic Words[67] | 8,400 |
| Attia et al. 2011c | Arabic Broken Plurals[68] | 2,562 |

Table 9: List of Words Lists.

### 4.4 Speech Corpora

To the best of our knowledge the corpus in Table 10 compiled by Almeman and Lee (2013) is the only freely available speech corpus for Arabic. Most of the currently available speech corpora are available from the LDC or ELRA with a membership fees.

| Source | Corpus | Files |
|---|---|---|
| Almeman and lee (2013) | Arabic Speech Corpora[69] | 67,132 |

Table 10: List of Speech Corpora.

### 4.5 Handwriting Recognition Corpora

Again, the handwriting recognition corpora are very rare in Arabic and they are mostly available at cost. The four corpora listed in Table 11 are an exception and they can be used for various NLP tasks from OCR to writer identification.

| Source | Corpus | Files |
|---|---|---|
| Al-Maadeed, et al (2011) | QUWI Handwritings Dataset[70] | 1,000 |
| Hassaïne and Maadeed (2012) | Writer Identification Contest for Arabic Scripts Data set[71] | 200 |
| Al-Maadeed, et al (2002) | AHDB Data Set[72] | 100 |
| Al-Maadeed, et al (2012) | ICDAR2011 competition Data set[73] | 50 |

Table 11: Handwriting Recognition Corpora.

### 4.6 Miscellaneous Corpora types

The list in Table 12 presents seven corpora useful for a multitude of NLP related tasks such as question answering Ben Ajiba et al. (2007) and Trigui et al. (2010), plagiarism detection Bensalem et al. (2013), document summarization El-Haj et al. 2010 and El-Haj and Rayson (2013), comparable text detection Saad et al. (2013).

Finally, the Kalimat multi-purpose corpus (El-Haj and Koulali (2013) is a unique corpus that includes around 20k newswire words extracted for summaries, named entities tagged, part of speech tagged and morphologically analyzed.

### 5. Conclusion

We presented the preliminary results of the first survey reserved for the freely Arabic Corpora. The goal behind this study is to promote the use of free corpora especially by those who lack funding and cannot afford membership or high fees to acquire a corpora from a language data center. The results obtained showed that many of the freely available resources for Arabic are not always visible and therefore it is hard be found by potential users.

---

53 http://catalog.ldc.upenn.edu/LDC2002L49
54 http://www.perseus.tufts.edu/hopper/opensource/download
55 http://www-igm.univ-mlv.fr/~unitex/index.php?page=3&htm
56 https://aralex.mrc-cbu.cam.ac.uk/aralex.online/login.jsp
57 http://sourceforge.net/projects/aracomlex/files/
58 http://ling.uni-konstanz.de/pages/home/mousser/files/Arabic_verbnet.php
59 http://sourceforge.net/projects/awnbrowser/
60 http://www.nooj4nlp.net/pages/arabic.html
61 http://sourceforge.net/projects/arabeyes/files/QaMoose/2.1/
62 http://arabicwordcount.sourceforge.net/
63 http://sourceforge.net/projects/arabic-wordlist/
64 https://sourceforge.net/projects/arabicmwes/
65 http://arabic-unknowns.sourceforge.net/
66 http://sourceforge.net/projects/arabicstopwords/
67 http://obsoletearabic.sourceforge.net/
68 http://broken-plurals.sourceforge.net/

69 http://www.cs.bham.ac.uk/~kaa846/arabic-speech-corpora.html
70 http://handwriting.qu.edu.qa/dataset/
71 http://handwriting.qu.edu.qa/dataset/
72 http://handwriting.qu.edu.qa/dataset/
73 http://handwriting.qu.edu.qa/dataset/

Moreover, the 66 corpora listed in this paper cover the main categories of corpora types. We hope that this initial attempt to located freely available Arabic corpora would be useful to the research community and such effort can be easily replicated to located similar sources for other languages. The corpora list presented in this paper, is made available in a single webpage for an easier access.[74] In the near future, we plan to make the list available in an online database and we will continue looking for other free corpora to enrich our repository.

| Source | Corpus | Words |
|---|---|---|
| Saad et al. (2013) | AFEWC and Enews Comparable Corpora[75] | 28,000,000 |
| Bensalem et al. (2013) | InAra (a corpus for Arabic Intrinsic plagiarism detection evaluation)[76] | 12,681,374 |
| El-Haj et al. (2010) | Essex Arabic Summaries Corpus[77] | 41,493 |
| El-Haj and Rayson (2013) | Multi-document Summaries[78] | 30,000 |
| El-Haj and Koulali (2013) | KALIMAT Multi-Purpose Corpus[79] | 20,291 |
| Ben Ajiba et al. (2007) | Arabic QA/IR[80] | 11,638 |
| Trigui et al. (2010) | Arabic Definition QA corpus[81] | 250 |

Table 12: Miscellaneous Corpora Types

---

[74] The current list of corpora is available at <http://www.qatar.cmu.edu/~wajdiz/corpora.html>

[75] http://sourceforge.net/projects/crlcl/

[76] https://sourceforge.net/projects/inaracorpus/

[77] http://sourceforge.net/projects/easc-corpus/

[78] http://multiling.iit.demokritos.gr/file/all

[79] http://sourceforge.net/projects/kalimat/

[80] http://www.dsic.upv.es/~ybenajiba/resources/Corpus.zip

[81] https://sites.google.com/site/anlprg/outils-et-corpus-realises/ArabicDefinitionQuestionAnsweringData.rar?attredirects=0